\documentclass[11pt]{article}

\usepackage[preprint]{acl}   %
\usepackage{times}
\usepackage{latexsym}
\usepackage[T1]{fontenc}
\usepackage[utf8]{inputenc}
\usepackage{microtype}
\usepackage{inconsolata}
\usepackage{graphicx}
\usepackage{amsmath}
\usepackage{amssymb}
\usepackage{array}
\usepackage{booktabs}
\usepackage{multirow}
\usepackage{tabularx}
\usepackage{enumitem}
\usepackage{xcolor}
\usepackage{colortbl}
\usepackage[most]{tcolorbox}
\usepackage{pifont}
\usepackage{makecell}
\tcbuselibrary{breakable}
\newtcolorbox{promptbox}[1]{
  width=\linewidth,
  breakable,
  colback=black!2,
  colframe=black!75,
  colbacktitle=black!75,
  coltitle=white,
fonttitle=\small\bfseries\sffamily,
fontupper=\footnotesize,
  title={#1},
  boxrule=0.4pt,
  arc=1.5mm,
  left=6pt,
  right=6pt,
  top=6pt,
  bottom=6pt,
  before skip=8pt,
  after skip=8pt
}

\definecolor{closedleader}{HTML}{ECECFE} %
\definecolor{openleader}{HTML}{E8F5EB}%
\definecolor{summaryhead}{HTML}{E8EDF3}
\definecolor{perceptionhead}{HTML}{E2F1F2}
\definecolor{understandinghead}{HTML}{E7EEF9}
\definecolor{reasoninghead}{HTML}{F9EEDC}
\definecolor{judgmenthead}{HTML}{F5E5ED}
\definecolor{takeawayframe}{HTML}{B65B49}
\definecolor{takeawaytitle}{HTML}{35735F}
\definecolor{relatedours}{RGB}{255,232,204} %

\newcommand{\dimensionhead}[4]{%
  \begingroup
  \setlength{\fboxsep}{0pt}%
  \shortstack[c]{%
    \raisebox{0.35ex}{\makebox[0pt][c]{\colorbox{#1}{%
      \makebox[#2][c]{\rule[-0.50ex]{0pt}{2.65ex}\textbf{#4}}}}}\\[0.20ex]
    \rule{#3}{0.30pt}}%
  \endgroup
}

\definecolor{asymwarmone}{RGB}{253,242,234}
\definecolor{asymwarmtwo}{RGB}{248,222,204}
\definecolor{asymwarmthree}{RGB}{243,199,170}
\definecolor{asymwarmfour}{RGB}{244,188,151}
\definecolor{asymwarmfive}{RGB}{240,176,134}
\definecolor{asymcoolone}{RGB}{238,245,248}
\definecolor{asymcooltwo}{RGB}{214,232,240}
\definecolor{asymcoolthree}{RGB}{185,216,230}
\definecolor{asymcoolfour}{RGB}{174,207,222}
\definecolor{asymcoolfive}{RGB}{160,198,216}
\definecolor{asymneutral}{RGB}{244,245,245}
\definecolor{asymmean}{RGB}{239,243,246}

\definecolor{closedleader}{HTML}{ECECFE}%
\definecolor{openleader}{HTML}{E8F5EB}%

\title{\textsc{Pure}: A Diagnostic Benchmark for Multi-Granularity Multimodal Identity Matching}

\author{
  Hengyu You,\quad Xintong Wang, \quad 
  Jin Cao, \quad 
  Liang Ding\thanks{Corresponding author.} \\
  Alibaba Group \\
  \texttt{\{youhengyu.yhy, hanfeng.wxt, luyu.cj, zuorui.dl\}@alibaba-inc.com}
}

\begin{document}

\maketitle
\begin{abstract}
Identity matching asks whether two records denote the same identity, yet the answer depends on the target granularity. Two records may match at the base level but not at the variant level: they share an underlying design while differing in attributes such as color or size. Final matching accuracy does not separately characterize the capabilities involved. To study this problem, we use e-commerce as a challenging real-world domain. We introduce \textbf{\textsc{Pure}}, a diagnostic benchmark for multi-granularity multimodal identity matching covering four capability dimensions: \textbf{P}erception, \textbf{U}nderstanding, \textbf{R}easoning, and \textbf{E}vidence-grounded Judgment. The benchmark comprises 1,457 multiple-choice questions across 14 tasks, derived from real-world matching pairs and verified by two domain experts. Across ten recent MLLMs, overall scores range from 47.1\% to 86.0\%. Understanding receives the highest mean score and Judgment the lowest, with weaknesses in visual difference localization and variant-level rule judgment. 
Modality ablations show that models benefit from multimodal evidence but can incorrectly identify conflicts between matching records. In direct matching, most models recognize matching pairs more reliably than non-matching pairs at the base level, whereas the reverse holds at the variant level.
 \textsc{Pure} therefore reveals what aggregate matching accuracy hides: not only whether models judge a pair correctly, but where they struggle to interpret evidence and apply matching criteria.
\end{abstract}

\section{Introduction}
\label{sec:introduction}

\begin{figure*}[t]
  \centering
  \includegraphics[width=\linewidth]{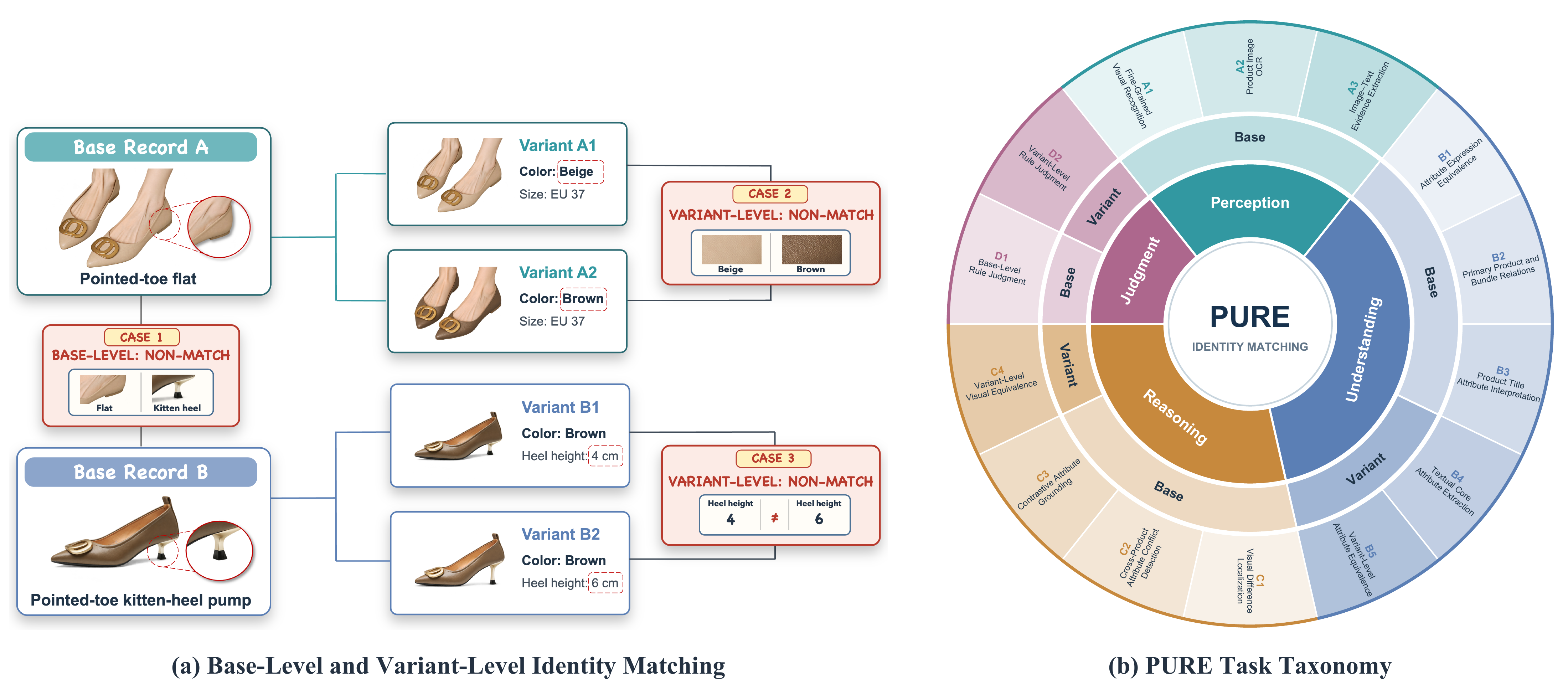} %
    \caption{\textbf{Multi-granularity identity matching and the \textsc{Pure} task taxonomy.}
(a) Records A and B differ in heel structure and do not match at the base level (Case~1). Within a shared base design, records that differ in color (Case~2) or heel height (Case~3) do not match at the variant level.
(b) \textsc{Pure} organizes 14 tasks by diagnostic dimension and identity-matching granularity.}
  \label{fig:motivation}
\end{figure*}

Identity matching asks whether two records denote the same underlying identity. Records may refer to the same identity despite differences in presentation, whereas superficially similar records may refer to different identities. This problem is studied both in human identity verification from faces and voices~\citep{smith2025facevoice} and in fine-grained object discrimination by vision-language models~\citep{marsili2026twin}. The latter shows that models can incorrectly match objects that share salient visual features but differ in local structure. We focus on multimodal identity matching, where each record describes an object through an image and text, and the model determines whether two records match.
Yet what counts as "the same" depends on the target granularity. %
Cognitive research shows that categorization rules modulate shape representations in the human visual cortex~\citep{henderson2025dynamic}. In computer vision, conditional image similarity likewise requires models to compare images according to specified attributes or objects~\citep{vaze2023genecis}. 

E-commerce provides a concrete and challenging setting for studying multi-granularity identity matching.
Differences in photography, title wording, and attribute expressions across sellers introduce substantial variation in presentation. %
We adopt two nested matching criteria. At the \textbf{\emph{base level}}, records must describe the same underlying design and agree on its core attributes, while differences are permitted in a specified set of additional attributes. At the \textbf{\emph{variant level}}, records must additionally agree on all attributes in this set. This matching granularity corresponds to the instance-level exact matching setting in Pailitao-VL~\citep{chen2026pailitaovl}. As illustrated in Figure~\ref{fig:motivation}(a), records A and B do not match at the base level because of the structural difference between flat and kitten-heel shoes (Case~1). Within a shared base design, records that differ in color (Case~2) or heel height (Case~3) do not match at the variant level. A variant-level match therefore implies a base-level match, but not conversely. These examples illustrate a central challenge in identity matching: \textbf{models must acquire and interpret multimodal evidence, compare attributes across records, and determine which differences should affect the matching decision at the target granularity.} 

Recent MLLM benchmarks assess visual perception, multi-image comparison, and comparative reasoning~\citep{fu2024blink,m4bench2025,vlmsubtlebench2026}. In e-commerce, benchmarks cover multimodal understanding, product matching, and rule-aware relevance assessment~\citep{ecommmmu2025,oxyecombench2026,rair2025}. %
However, in direct identity matching, final accuracy does not separately characterize model performance on the capabilities involved.

We therefore introduce \textbf{\textsc{Pure}}\footnote{The dataset is available at \url{https://huggingface.co/datasets/alibaba-multimodal-industrial-ai/PURE}, and the evaluation code at \url{https://github.com/alibaba-multimodal-industrial-ai/PURE}.}, a diagnostic benchmark for multi-granularity multimodal identity matching, comprising 1,457 multiple-choice questions across 14 tasks in four dimensions: Perception, Understanding, Reasoning, and Judgment. The questions are derived from real-world e-commerce pairs and verified by two domain experts.
Across ten MLLMs, Understanding is strongest on average, whereas Judgment is weakest. 
Modality ablations reveal task-specific evidence dependence and suggest that models can mistake non-discriminative textual differences for identity conflicts.
In direct matching, most models achieve higher recall on matching pairs than on non-matching pairs at the base level, whereas the reverse holds at the variant level. Overall \textsc{Pure} Scores also correlate with performance on direct identity matching and two independently constructed benchmarks.
Together, these findings show that \textsc{Pure} complements aggregate matching accuracy by revealing capability gaps, dependence on input evidence, and distinct patterns across matching granularities. 
Our contributions are as follows: 

\begin{itemize}[leftmargin=*, labelsep=0.5em]
    \item \textbf{A multi-granularity diagnostic benchmark.} We formalize multimodal identity matching under base-level and variant-level criteria, defining 14 tasks across four capability dimensions (\S\ref{sec:benchmark}).

  \item \textbf{An error-driven construction pipeline.} We construct questions from misclassified real-world record pairs, combining task assignment, multi-model screening, and verification by two domain experts (\S\ref{sec:construction}).

    \item \textbf{Empirical findings on identity matching.} Experiments with ten MLLMs reveal a gap between attribute understanding and identity judgment, task-specific evidence dependence, and matching asymmetries across granularities (\S\ref{sec:experiments}).

\end{itemize}

\section{Related Work}
\label{sec:related}

\begin{table*}[t]
  \centering
  \caption{\textbf{Comparison with closely related benchmarks.} Evidence modalities indicate the modalities required for models to answer questions. Identity matching indicates whether at least one task explicitly evaluates identity matching.
Multi-granularity matching indicates whether identity matching is evaluated at multiple levels of granularity.}
  \small
  \setlength{\tabcolsep}{6pt}
  \renewcommand{\arraystretch}{1.15}
  \renewcommand{\tabularxcolumn}[1]{m{#1}}
  \begin{tabular*}{\textwidth}{@{\extracolsep{\fill}}llccl@{}}
    \toprule
    \textbf{Benchmark} & \makecell[l]{\textbf{Evidence}\\\textbf{modalities}} & \makecell[c]{\textbf{Identity}\\\textbf{matching}} & \makecell[c]{\textbf{Multi-granularity}\\\textbf{matching}} & \textbf{Evaluation focus} \\ %
    \midrule
    M4Bench~\citep{m4bench2025} & Images & $\checkmark$ & $\times$ & Visual comparison \\
    FGVQA~\citep{marsili2026twin} & Images & $\checkmark$ & $\times$ & Fine-grained recognition \\
    VLM-SubtleBench~\citep{vlmsubtlebench2026} & Images & $\times$ & $\times$ & Subtle comparative reasoning \\
    EcomMMMU~\citep{ecommmmu2025} & Images, text & $\times$ & $\times$ & E-commerce understanding \\
    RAIR~\citep{rair2025} & Images, text & $\times$ & $\times$ & Search relevance \\
    OxyEcomBench~\citep{oxyecombench2026} & Images, text & $\checkmark$ & $\times$ & Broad e-commerce tasks \\
    \midrule
    \textbf{\textsc{Pure}} & Images, text & $\checkmark$ & $\checkmark$ & Capabilities for identity matching \\  %
    \bottomrule
  \end{tabular*}
  \label{tab:related-benchmarks}
\end{table*}

\paragraph{Fine-grained visual comparison and identity matching.} 
Fine-grained visual comparison examines whether models can identify subtle differences and correspondences across images.
Product1M, M5Product, and FGPR study multimodal product representations and retrieval~\citep{zhan2021product1m,dong2022m5product,zhang2026fgpr}.
Pailitao-VL evaluates exact matching at a granularity corresponding to our variant level~\citep{chen2026pailitaovl}.
BLINK and MIBench assess visual perception and multi-image understanding~\citep{fu2024blink,liu2024mibench}, while MLLM-CompBench, M4Bench, and VLM-SubtleBench evaluate cross-image comparison and fine-grained discrimination~\citep{kil2024mllmcompbench,m4bench2025,vlmsubtlebench2026}.
FGVQA evaluates fine-grained recognition across six datasets~\citep{marsili2026twin}. NaturalBench pairs each question with two images that require different answers, assessing whether models use visual evidence rather than rely on language priors~\citep{naturalbench2024}.
However, these visual comparison benchmarks do not evaluate identity matching under separate base-level and variant-level criteria.

\paragraph{\mbox{E-commerce benchmarks.}}
Existing benchmarks in e-commerce cover diverse tasks, including product understanding, matching, and search relevance.
Shopping MMLU evaluates textual shopping skills~\citep{jin2024shoppingmmlu}, while EcomMMMU and OxyEcomBench cover diverse multimodal e-commerce tasks~\citep{ecommmmu2025,oxyecombench2026}.
For industrial applications, IndustryBench evaluates LLMs on multilingual industrial procurement question answering~\citep{industrybench2026}. IndustryBench-MIPU evaluates MLLMs on attribute value extraction from multiple industrial product images~\citep{industrybenchmipu2026}.
RAIR evaluates rule-aware search relevance~\citep{rair2025}. 
TaoSR1 includes an atomic capability benchmark covering four dimensions of relevance modeling~\citep{taosr12025}. 
However, identity matching is not a central focus of these benchmarks. When included, it is evaluated as one of several application tasks, without a dedicated assessment of the capabilities supporting identity matching.

Table~\ref{tab:related-benchmarks} compares the scope and focus of closely related benchmarks.
\textsc{Pure} assesses evidence acquisition, attribute understanding, cross-record comparison, and evidence-grounded judgment for identity matching at the base and variant levels, providing capability profiles that complement final matching results.

\section{Methodology}
\label{sec:methodology}
We define identity matching at the base and variant levels and design 14 evaluation tasks across four capability dimensions (Section~\ref{sec:benchmark}). We then describe how questions are constructed from real-world matching errors, screened, verified, and assembled into the final benchmark (Section~\ref{sec:construction}).

\subsection{PURE Framework}
\label{sec:benchmark}
Matching granularity specifies which properties must agree for two records to match. The task taxonomy organizes the assessment into four capability dimensions: Perception, Understanding, Reasoning, and Judgment. We instantiate these tasks with e-commerce records containing images, titles, and attribute descriptions.

\subsubsection{Base-Level and Variant-Level Identity Matching}
\label{sec:formulation}

We study identity matching over two multimodal records, each of which may contain images, descriptive text, and structured attributes. We define two nested matching criteria that differ in what counts as the same identity rather than in the input format.

Two records match at the \textbf{base level} when they refer to the same underlying design: their identity-defining structure, composition, and core attributes are equivalent, while differences confined to a specified set of additional attributes are permitted. Presentation differences such as viewpoint, background, seller-specific wording, and marketing layout do not change their base-level identity. Two records match at the \textbf{variant level} when they match at the base level and additionally agree on all applicable attributes in this set. Writing $a \sim_g b$ for a match between records $a$ and $b$ at granularity $g$, we have
\begin{equation}
  a \sim_{\mathrm{variant}} b
  \;\Longrightarrow\;
  a \sim_{\mathrm{base}} b,
  \label{eq:type-instance-relation}
\end{equation}
whereas the converse need not hold: two records may match at the base level but differ at the variant level.
For example, two listings of the same shoe design in different colors can match at the base level but differ at the variant level. Listings from different sellers can still match at the variant level when they describe the same design and agree on all specified attributes despite differences in imagery or wording. %

Identity-matching systems typically produce either a pairwise same/different prediction or a similarity score used for retrieval. These outputs characterize the final matching outcome but do not show how well a model performs the individual tasks involved. A matching error may involve a missed visual detail, a misunderstood attribute expression, an incorrect cross-record comparison, or the misapplication of an identity rule. \textsc{Pure} therefore complements direct matching evaluation with task-specific questions that assess these capabilities separately.

Each \textsc{Pure} question follows a multiple-choice format and consists of task-specific evidence, a question stem, answer options, and a reference answer. Depending on the task, the model-visible evidence is drawn from listing images, titles, and descriptions of the selected variant's attributes.

\subsubsection{Task Taxonomy}
\label{sec:taxonomy}

\textsc{Pure} comprises 14 tasks across two matching granularities, organized into four dimensions: Perception, Understanding, Reasoning, and Judgment. These dimensions assess evidence acquisition, semantic interpretation and normalization, comparative reasoning and consistency, and the application of identity criteria, respectively.

\paragraph{Perception.}
Perception examines whether a model can acquire identity-relevant evidence before comparing products. \emph{Fine-Grained Visual Recognition} (A1) identifies local appearance and structural cues, \emph{Product Image OCR} (A2) reads text printed on products and packaging, and \emph{Image--Text Evidence Extraction} (A3) extracts identity-relevant attributes from a product image and its corresponding title.

\paragraph{Understanding.}
Understanding examines whether product evidence can be interpreted and normalized. \emph{Attribute Expression Equivalence} (B1) identifies semantically equivalent attribute expressions; \emph{Primary Product and Bundle Relations} (B2) distinguishes the primary product from accessories and bundle components; and \emph{Product Title Attribute Interpretation} (B3) resolves product terminology, model expressions, and title attributes. At the variant level, \emph{Textual Core Attribute Extraction} (B4) identifies the selected variant's core attributes from the title and attribute text, resolving ambiguity or conflicting descriptions. \emph{Variant-Level Attribute Equivalence} (B5) determines whether two selected variants have semantically equivalent attribute descriptions despite surface-form differences.

\paragraph{Reasoning.}
Reasoning examines differences and consistency across product evidence. \emph{Visual Difference Localization} (C1) identifies or rules out identity-relevant visual differences, while \emph{Cross-Product Attribute Conflict Detection} (C2) detects identity-relevant attribute conflicts between two product listings. \emph{Contrastive Attribute Grounding} (C3) evaluates two similar products against two competing attribute descriptions as a four-question group. \emph{Variant-Level Visual Equivalence} (C4) determines whether two images depict the same identity at the variant level by distinguishing identity-relevant visual differences from changes in presentation.

\begin{figure*}[t]
  \centering
  \includegraphics[width=\textwidth]{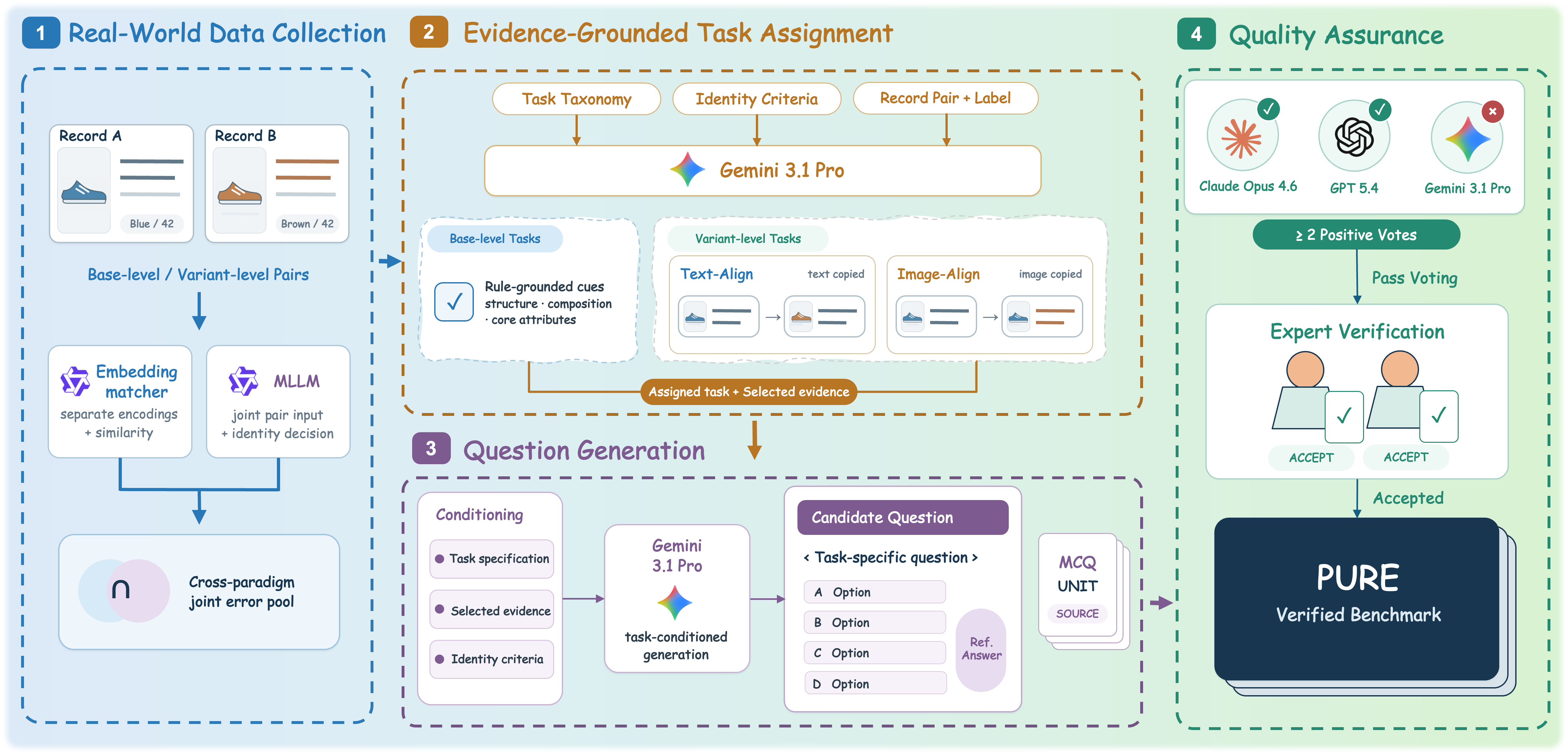} %
    \caption{\textbf{Construction pipeline of \textsc{Pure}.} \ding{182}~Real-world base-level and variant-level record pairs jointly misclassified by an embedding matcher and a generative MLLM form the source pool. \ding{183}~Identity criteria and the task taxonomy guide evidence-grounded analysis and task assignment. \ding{184}~Gemini 3.1 Pro generates task-conditioned questions. \ding{185}~Candidates undergo multi-model quality screening followed by expert review.}
  \label{fig:construction}
\end{figure*}

\paragraph{Judgment.}
Judgment examines whether a model can apply identity criteria to identify the non-match rules supported by a record pair. \emph{Base-Level Rule Judgment} (D1) asks which base-level non-match rule, if any, the pair triggers. \emph{Variant-Level Rule Judgment} (D2) identifies which variant-level non-match rule, if any, a record pair triggers. Both tasks include an option indicating that no rule applies, allowing the model to distinguish pairs with a supported conflict from those that satisfy the matching criterion.

The task taxonomy is developed through human analysis of identity rules and recurrent matching errors. 
The complete taxonomy is reported in Table~\ref{tab:taxonomy} in Appendix~\ref{app:task-definitions}, including each task's name, dimension-aligned identifiers, target identity-matching granularity, evidence configurations, and primary diagnostic focus. Figure~\ref{fig:motivation}(b) visualizes its organization across dimension and granularity. %

\subsection{Benchmark Construction}
\label{sec:construction}

Having defined the target identity-matching decision and its tasks, we construct \textsc{Pure} through four stages: \ding{182}~\emph{Real-World Data Collection}, \ding{183}~\emph{Evidence-Grounded Task Assignment}, \ding{184}~\emph{Question Generation}, and \ding{185}~\emph{Quality Assurance}.

\subsubsection{Real-World Data Collection}
\label{sec:candidates}

We begin with real-world e-commerce record pairs labeled under base-level or variant-level matching criteria. To identify challenging cases, each pair is evaluated by two distinct matching paradigms. An embedding-based matcher encodes the records independently and derives a decision from their representation similarity, whereas a generative MLLM jointly observes both records and predicts whether they match at the designated granularity. We retain pairs for which both matchers disagree with the reference label at that granularity. 
The retained pool covers false-positive and false-negative cases at both levels. This intersection emphasizes errors shared by two substantially different matching mechanisms. The resulting pairs provide source material for constructing questions about the evidence and capabilities involved in these difficult decisions.

\begin{table*}[t]
  \centering
  \caption{\textbf{Overall, dimension-level, and task-level results on \textsc{Pure} (\%).} Within each diagnostic dimension, task scores are followed by the corresponding dimension score (\textbf{Total}), and thin vertical rules separate dimensions. Models are ranked by overall score. The leading closed- and open-source models are shaded in {\setlength{\fboxsep}{2pt}%
\colorbox{closedleader}{\vphantom{purplegreen}purple} and
\colorbox{openleader}{\vphantom{purplegreen}green}}, respectively. C3 uses four-response group accuracy, and all other tasks use question-level accuracy. \textbf{Bold} and \underline{underlined} values mark the best and second-best metric scores, respectively. } %
  \label{tab:main-consolidated-grouped-candidate}
  \scriptsize
  \setlength{\tabcolsep}{2.25pt} %
  \renewcommand{\arraystretch}{1.08}
  \begin{tabularx}{\textwidth}{@{\hspace{4pt}}>{\raggedright\arraybackslash}Xc
  !{\color{black!28}\vrule width 0.35pt}
    r
    !{\color{black!28}\vrule width 0.35pt}
    *{3}{r}>{\columncolor{perceptionhead!55!white}}r
    !{\color{black!28}\vrule width 0.35pt}
    *{5}{r}>{\columncolor{understandinghead!55!white}}r
    !{\color{black!28}\vrule width 0.35pt}
    *{4}{r}>{\columncolor{reasoninghead!55!white}}r
    !{\color{black!28}\vrule width 0.35pt}
    *{2}{r}>{\columncolor{judgmenthead!55!white}}r}
    \toprule
 \multirow[c]{2}{*}{\raisebox{1.0ex}{\textbf{Model}}}
    & \multirow[c]{2}{*}{\raisebox{1.0ex}{\textbf{Rank}}}
    & \multicolumn{1}{c!{\color{black!28}\vrule width 0.35pt}}{\multirow[c]{2}{*}{\raisebox{1.0ex}{\textbf{Overall}}}}
      & \multicolumn{4}{c}{\dimensionhead{perceptionhead}{2.55cm}{2.35cm}{Perception}}
    & \multicolumn{6}{c}{\dimensionhead{understandinghead}{3.80cm}{3.60cm}{Understanding}}
    & \multicolumn{5}{c}{\dimensionhead{reasoninghead}{3.20cm}{3.00cm}{Reasoning}}
    & \multicolumn{3}{c}{\dimensionhead{judgmenthead}{1.90cm}{1.70cm}{Judgment}} \\
    & &
    & \textbf{A1} & \textbf{A2} & \textbf{A3} & \textbf{Total}
    & \textbf{B1} & \textbf{B2} & \textbf{B3} & \textbf{B4} & \textbf{B5} & \textbf{Total}
    & \textbf{C1} & \textbf{C2} & \textbf{C3} & \textbf{C4} & \textbf{Total}
    & \textbf{D1} & \textbf{D2} & \textbf{Total} \\
    \midrule
    \rowcolor{black!7}
    \multicolumn{21}{l}{\textit{Closed-source models}} \\
    \rowcolor{closedleader}[4pt][2.3pt]%
    Gemini 3.8 Flash
      & (1) & \textbf{86.0}
      & \underline{79.6} & \textbf{91.9} & \textbf{82.7} & \textbf{84.7}
      & \textbf{97.3} & 96.1 & \textbf{94.7} & \underline{97.0} & \textbf{99.0} & \textbf{96.8}
      & \textbf{81.7} & \textbf{82.9} & \underline{66.7} & 91.0 & \textbf{80.6}
      & \textbf{75.5} & \textbf{68.0} & \textbf{71.8} \\
    GPT-5.6 Sol
      & (2) & \underline{79.7}
      & 65.6 & 65.9 & 76.9 & 69.5
      & \underline{86.4} & \textbf{99.0} & \underline{92.0} & \textbf{98.0} & \textbf{99.0} & \underline{94.9}
      & \underline{80.3} & 70.7 & \underline{66.7} & 92.3 & \underline{77.5}
      & 68.3 & 54.4 & 61.4 \\
    Qwen3.8-Max
      & (3) & 77.0
      & \textbf{87.1} & 64.2 & 71.2 & \underline{74.2}
      & 65.8 & 93.1 & 88.0 & 96.0 & \underline{97.0} & 88.0
      & 47.9 & 63.4 & \textbf{76.7} & \textbf{96.2} & 71.0
      & \underline{73.4} & 57.6 & 65.5 \\
    Claude Opus 5
      & (4) & 75.7
      & 65.6 & \underline{78.9} & 65.4 & 69.9
      & 66.8 & \underline{97.1} & 85.3 & 96.0 & \underline{97.0} & 88.5
      & 59.2 & 63.4 & 60.0 & \underline{93.6} & 69.0
      & 72.7 & \underline{59.2} & \underline{65.9} \\
    \midrule
    \rowcolor{black!7}
    \multicolumn{21}{l}{\textit{Open-source models}} \\
    \rowcolor{openleader}[4pt][2.3pt]%
    Kimi K3
      & (5) & 73.1
      & 65.6 & 66.7 & \underline{81.7} & 71.3
      & 68.5 & 96.1 & 88.0 & \textbf{98.0} & \textbf{99.0} & 89.9
      & 38.0 & 56.1 & 53.3 & 89.7 & 59.3
      & \underline{73.4} & 48.8 & 61.1 \\
    DeepSeek-V4.1-Flash
      & (6) & 69.6
      & 71.0 & 48.0 & 72.1 & 63.7
      & 62.0 & 92.2 & 77.3 & 96.0 & 95.0 & 84.5
      & 32.4 & \underline{80.5} & 60.0 & 84.6 & 64.4
      & 67.6 & 36.0 & 51.8 \\
    Qwen3.5-397B-A17B
      & (7) & 59.2
      & 69.9 & 48.8 & 43.3 & 54.0
      & 46.7 & 85.3 & 61.3 & \underline{97.0} & \underline{97.0} & 77.5
      & 39.4 & 26.8 & 50.0 & 80.8 & 49.3
      & 45.3 & 37.6 & 41.5 \\
    Qwen3.5-27B
      & (8) & 58.2
      & 51.6 & 52.0 & 55.8 & 53.1
      & 35.3 & 81.4 & 53.3 & \textbf{98.0} & \underline{97.0} & 73.0
      & 25.4 & 61.0 & 33.3 & 83.3 & 50.7
      & 51.1 & 36.8 & 43.9 \\
    Qwen3.5-35B-A3B
      & (9) & 51.0
      & 36.6 & 42.3 & 49.0 & 42.6
      & 27.7 & 79.4 & 38.7 & 95.0 & 84.2 & 65.0
      & 9.9 & 43.9 & 50.0 & 83.3 & 46.8
      & 38.8 & 35.2 & 37.0 \\
      Qwen3.5-9B
      & (10) & 47.1
      & 51.6 & 29.3 & 37.5 & 39.5
      & 31.5 & 68.6 & 18.7 & 89.1 & 89.1 & 59.4
      & 22.5 & 36.6 & 26.7 & 76.9 & 40.7
      & 45.3 & 36.0 & 40.7 \\
    \bottomrule
    \end{tabularx}
\end{table*}

\subsubsection{Evidence-Grounded Task Assignment}
\label{sec:assignment}

Each retained pair is assigned to one predefined task before question generation. As illustrated in Figure~\ref{fig:construction}, Gemini 3.1 Pro~\citep{googledeepmind2026gemini31pro} receives the original record evidence, the reference matching label at the designated granularity, the two matcher outputs, the applicable identity criteria, and the task definitions. It produces a structured assignment record that identifies the evidence supporting the reference decision and recommends the task most relevant to interpreting that evidence correctly. The assignment record and selected evidence are retained for question generation and subsequent expert verification.

For variant-level pairs labeled as matches in the retained pool, which the generative matcher incorrectly predicts as non-matches, we additionally use evidence-alignment probes to inform task assignment. 
\emph{Text-Align} replaces record B's title and the attribute text of its selected variant with the corresponding text from record A while preserving both images. Conversely, \emph{Image-Align} replaces B's image with A's while preserving the original text. We rerun the same generative matcher on each modified pair and record whether its prediction now agrees with the reference matching label. A correction under Text-Align but not Image-Align supports assignment to Textual Core Attribute Extraction (B4). A correction under Image-Align but not Text-Align supports Variant-Level Visual Equivalence (C4). These probes guide task assignment based on the matcher's responses to changes in textual and visual evidence.

\subsubsection{Question Generation}
\label{sec:generation}

For each assigned case, Gemini 3.1 Pro receives the target task specification, applicable identity criteria, selected source evidence, and the assignment record. It generates a multiple-choice question with its answer options and reference answer. Questions in product image OCR (A2) and image--text evidence extraction (A3) allow multiple selections, whereas the other tasks require a single selected option. Inspired by the grouped evaluation design of NaturalBench~\citep{naturalbench2024}, the contrastive attribute grounding task (C3) combines two related records with two discriminative attribute statements. Each record is queried about both statements, yielding four questions whose responses jointly constitute one scoring unit. This structure assesses consistent attribute grounding across the two records.

\subsubsection{Quality Assurance}
\label{sec:automatic-qc}
\label{sec:expert-review}

Quality assurance has two stages. First, Claude Opus 4.6~\citep{anthropic2026claudeopus46}, Gemini 3.1 Pro, and GPT-5.4~\citep{openai2026gpt54} independently screen each candidate for evidence sufficiency, option completeness, unique answer support, and consistency with its assigned task and identity granularity. A candidate proceeds only when at least two models accept it. This stage retains approximately 72\% of roughly 3,940 candidate questions, yielding 2,819 questions for expert review. %

Second, two qualified e-commerce domain experts review every surviving unit against the source evidence and applicable identity criteria. Units that fail any review criterion are removed, and disagreements are resolved by jointly re-examining the source evidence. A unit is retained only when both experts approve it. %

Following expert review and final task-wise assembly, the benchmark contains 1,457 questions corresponding to 1,367 scoring units, including 30 complete C3 groups. %
Reviewer qualifications, detailed instructions, and count definitions are provided in Appendix~\ref{app:human-review}.

\section{Experiments}
\label{sec:experiments}
We evaluate \textsc{Pure} through three research questions.  RQ1 characterizes MLLM performance across tasks and capability dimensions. %
RQ2 tests models' dependence on visual and textual evidence. RQ3 examines whether \textsc{Pure} scores align with direct identity matching and independently constructed external benchmarks, while characterizing granularity-dependent behavior in direct matching. %

\subsection{Experiment Setup}
\label{sec:setup}

\subsubsection{Baselines}

We evaluate ten recent MLLMs, including closed-source models and open-source models of different sizes. The \textbf{closed-source} baselines are Claude Opus 5~\citep{anthropic2026claudeopus5}, Gemini 3.8 Flash~\citep{googledeepmind2026gemini38flash}, GPT-5.6 Sol~\citep{openai2026gpt56}, and Qwen3.8-Max~\citep{qwen2026qwen38}. The \textbf{open-source} baselines are four Qwen3.5 variants (9B, 27B, 35B-A3B, and 397B-A17B)~\citep{qwen2026qwen35}, DeepSeek-V4.1-Flash~\citep{deepseekai2026deepseekv41flash}, and Kimi K3~\citep{kimiteam2026kimik3}. %

\subsubsection{Evaluation Protocols and Metrics} %
\label{sec:metrics}

All models are evaluated under a common zero-shot protocol. For each question, the model receives the question stem and answer options together with the task-specific evidence defined by the taxonomy, drawn from listing images, titles, and descriptions of the selected variant's attributes as applicable. Task instructions and the required output format remain fixed across models. Each model is instructed to return only a JSON object whose \texttt{answer} field contains the selected option letter or letters. \footnote{The main comparison primarily uses non-thinking configurations. Paired comparisons with and without thinking are reported in Appendix~\ref{app:thinking}.} %

For each task, we report exact-match accuracy as the task-level metric. C3 is evaluated using group accuracy (\(\mathrm{G\text{-}Acc}\)), following NaturalBench~\citep{naturalbench2024}. A group is correct only when all four constituent questions are answered correctly. Each dimension score is the average of its constituent task scores, and the overall score is the average across all 14 tasks. %
Formal definitions are provided in Appendix~\ref{app:metric-definitions}. The evaluation prompts are provided in Appendix~\ref{app:prompts}.

\subsection{Capability Profiles (RQ1)}
\label{sec:rq1}

\paragraph{Overall performance.} %
Table~\ref{tab:main-consolidated-grouped-candidate} consolidates the overall, dimension-level, and task-level results. Overall scores range from 47.1\% to 86.0\%, revealing substantial variation across the evaluated MLLMs.
Gemini 3.8 Flash ranks first and leads all four dimensions, and Kimi K3 ranks first among the open-source models. %

\paragraph{Strengths and weaknesses.}
Understanding is the strongest dimension for every evaluated model, whereas Judgment is the weakest for most. Models perform consistently well on textual core attribute extraction (B4) and attribute equivalence (B5), but remain less reliable at locating visual differences (C1) and identifying applicable rules at the variant level (D2). These results reveal a gap between understanding attributes and identifying the visual evidence and rules that determine identity.

\paragraph{Scaling behavior.}
The Qwen3.5 series includes dense and mixture-of-experts variants at different parameter scales, with overall scores ranging from 47.1\% to 59.2\%. The dense 27B model nearly matches 397B-A17B and outperforms 35B-A3B, while 9B still exceeds 35B-A3B on several individual tasks. Moreover, the gains of 27B and 397B-A17B over 9B and 35B-A3B are larger in Perception and Understanding than in Reasoning and Judgment. Additional capacity is therefore associated more consistently with acquiring and interpreting evidence, while applying identity criteria remains a shared weakness. %

\begin{tcolorbox}[
  colback=black!4,
  colframe=black!50,
  coltext=black,
  boxrule=0.5pt,
  arc=2mm,
  left=6pt,right=6pt,
  top=5pt,bottom=5pt,
  before skip=6pt,after skip=6pt,
  fontupper=\itshape
]
\textcolor{takeawaytitle}{\textbf{Takeaway:}}
Models are more accurate at extracting and interpreting attributes than at localizing visual differences or applying identity criteria.

\end{tcolorbox}

\subsection{Evidence Dependence under Modality Ablation (RQ2)}
\label{sec:rq2}

We conduct paired modality ablations on A3, C2, and D1, covering evidence extraction, cross-record comparison, and rule judgment. Each question is evaluated under three conditions: full input, without listing images, and without listing titles. Task instructions and answer options remain fixed across conditions. We define $\Delta_m=\mathrm{Acc}_{\mathrm{Full}}-\mathrm{Acc}_{\mathrm{Ablated}}$, where $m\in\{\mathrm{image},\mathrm{title}\}$ and a positive value indicates a decrease in accuracy after evidence removal. These ablations assess how models' predictive behavior depends on the supplied evidence.  %

\begin{figure}[t]
  \centering
  \includegraphics[width=\columnwidth]{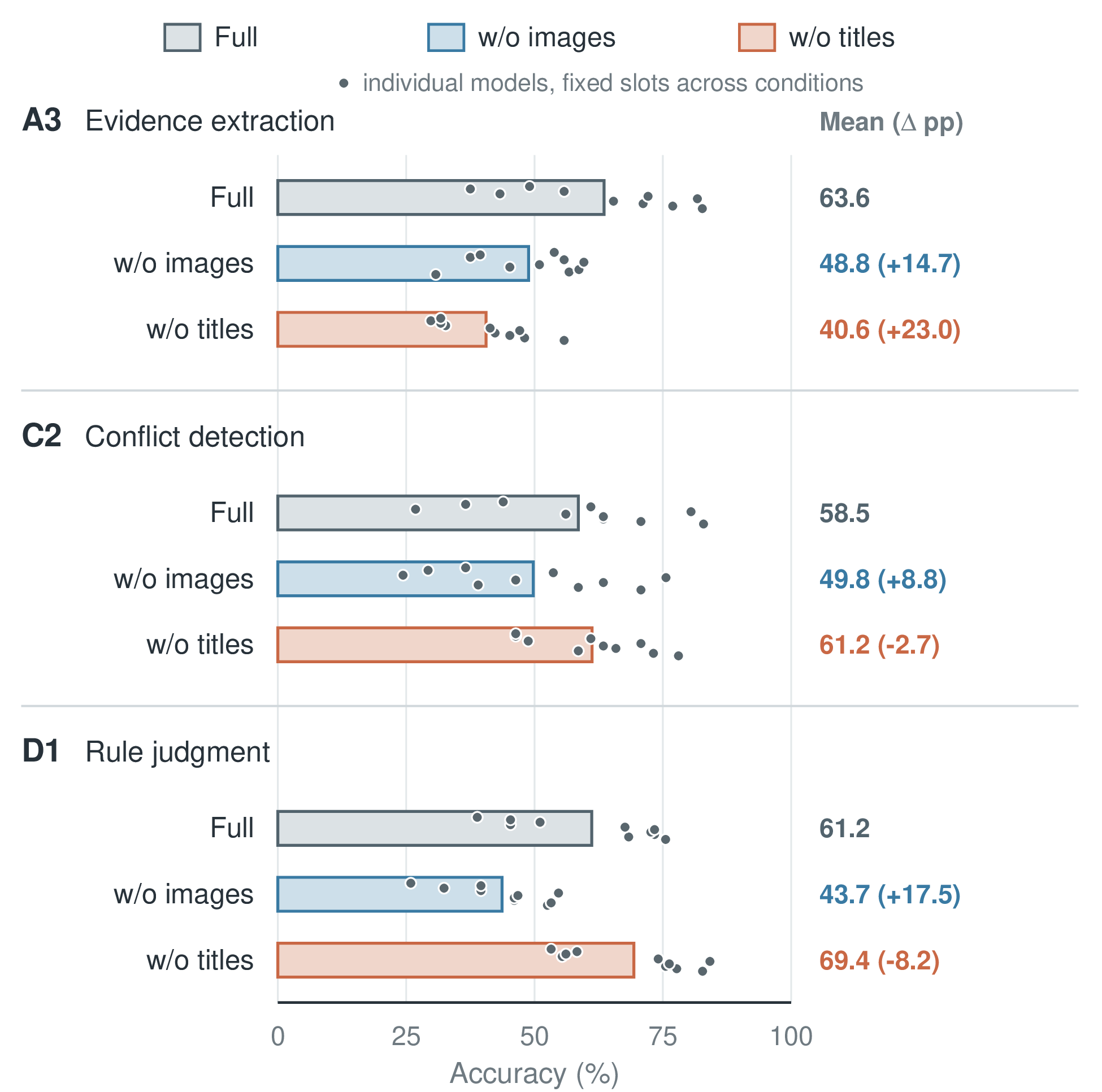}
  \caption{\textbf{Modality-ablation results across ten MLLMs.} Bars show unweighted mean accuracy under the full-input and two ablated conditions, and gray dots denote individual models. Parenthetical values report $\Delta_m=\mathrm{Acc}_{\mathrm{Full}}-\mathrm{Acc}_{\mathrm{Ablated}}$ in percentage points.}
  \label{fig:rq2-modality-ablation}
\end{figure}

Figure~\ref{fig:rq2-modality-ablation} summarizes the ablation results across all ten models. As expected, removing either the image or the title reduces mean accuracy on A3. In C2, removing images reduces mean accuracy by 8.8 percentage points, whereas removing titles increases it by 2.7 points, with six of the ten models showing higher accuracy. In D1, removing images reduces mean accuracy by 17.5 points, whereas removing titles increases it by 8.2 points, and this pattern holds for all ten models.
Further analysis shows that removing listing titles increases accuracy when the correct answer is "no applicable rule" but decreases it when the correct answer specifies a non-match rule (Appendix~\ref{app:rq2-details}). These effects suggest that textual evidence helps models identify valid non-match rules but can also induce false conflict judgments on matching pairs. This pattern echoes the "blind faith in text" observed under controlled cross-modal conflicts~\citep{deng2025words}.

\begin{tcolorbox}[
  colback=black!4,
  colframe=black!50,
  coltext=black,
  boxrule=0.5pt,
  arc=2mm,
  left=6pt,right=6pt,
  top=5pt,bottom=5pt,
  before skip=6pt,after skip=6pt,
  fontupper=\itshape
]
\textcolor{takeawaytitle}{\textbf{Takeaway:}}
Current MLLMs benefit from multimodal evidence, yet sometimes mistake non-discriminative textual differences for identity conflicts.  %
\end{tcolorbox}

\subsection{Diagnostic Relevance (RQ3)}  %
\label{sec:rq3}

\subsubsection{Relation to Direct Base-Level and Variant-Level Matching}
To examine the relationship between performance on \textsc{Pure} and direct identity matching, we reuse record pairs from \textsc{Pure} tasks involving two products. The same ten MLLMs directly predict binary labels, judging whether the two products match at the corresponding base or variant level. The evaluation set contains 330 base-level pairs and 211 variant-level pairs, with balanced class distributions, respectively. We report balanced accuracy and class-specific recall in Appendix~\ref{app:direct-matching}. %
Across the ten models, overall scores on \textsc{Pure} show a strong Spearman rank correlation~\citep{spearman1904proof} with direct-matching balanced accuracy ($\boldsymbol{\rho=0.952}$).

\begin{table}[t]
  \centering
  \caption{\textbf{Recall asymmetry in direct identity matching.} Gaps are $R_{\mathrm{M}}-R_{\mathrm{NM}}$ in percentage points, where $R_{\mathrm{M}}$ and $R_{\mathrm{NM}}$ denote recall on matching and non-matching pairs, respectively. Positive values indicate higher match recall, and negative values indicate higher non-match recall. The final row reports the mean across ten models.}
  \label{tab:direct-relation}
  \footnotesize
  \setlength{\tabcolsep}{4pt}
  \renewcommand{\arraystretch}{1.03}
    \begin{tabular}{@{}lcc}
    \toprule
    \textbf{Model} & \textbf{Base Gap} & \textbf{Variant Gap} \\
    \midrule
    Gemini 3.8 Flash
      & \cellcolor{asymwarmthree}$+20.0$
      & \cellcolor{asymcoolfive}$-37.2$ \\
    GPT-5.6 Sol
      & \cellcolor{asymneutral}$-2.4$
      & \cellcolor{asymcoolfive}$-42.1$ \\
    Qwen3.8-Max
      & \cellcolor{asymwarmfour}$+26.1$
      & \cellcolor{asymcoolthree}$-22.1$ \\
    Claude Opus 5
      & \cellcolor{asymneutral}$+2.4$
      & \cellcolor{asymcooltwo}$-9.2$ \\
    Kimi K3
      & \cellcolor{asymwarmthree}$+18.2$
      & \cellcolor{asymcoolone}$-6.6$ \\
    DeepSeek-V4.1-Flash
      & \cellcolor{asymneutral}$-1.8$
      & \cellcolor{asymcoolfive}$-36.0$ \\
    Qwen3.5-397B-A17B
      & \cellcolor{asymwarmtwo}$+8.5$
      & \cellcolor{asymcoolthree}$-16.3$ \\
    Qwen3.5-27B
      & \cellcolor{asymwarmfive}$+33.9$
      & \cellcolor{asymcoolone}$-5.1$ \\
    Qwen3.5-35B-A3B
      & \cellcolor{asymwarmfive}$+39.4$
      & \cellcolor{asymwarmthree}$+18.4$ \\
      Qwen3.5-9B
      & \cellcolor{asymwarmfour}$+27.3$
      & \cellcolor{asymneutral}$+2.2$ \\
    \midrule
    \textbf{Mean}
       & \cellcolor{asymwarmthree}\textbf{$+17.2$}
      & \cellcolor{asymcoolthree}\textbf{$-15.4$} \\
    \bottomrule
  \end{tabular}
\end{table}

Table~\ref{tab:direct-relation} shows different patterns at the two identity levels. At the base level, eight of ten models achieve higher recall on matching pairs than on non-matching pairs. At the variant level, eight show higher recall on non-matching pairs. The mean recall gap ($R_{\mathrm{M}}-R_{\mathrm{NM}}$) changes from $+17.2$ to $-15.4$ percentage points. 
\textbf{Most models recognize matching pairs more reliably than non-matching pairs at the base level, whereas the reverse holds at the variant level.}

\subsubsection{Relation to External Benchmarks} %

Motivated by VLM-SubtleBench's model-level relevance analysis~\citep{vlmsubtlebench2026}, we compare \textsc{Pure} with two closely related external benchmarks. M4Bench~\citep{m4bench2025} includes multi-image product comparison tasks relevant to e-commerce matching, whereas FGVQA~\citep{marsili2026twin} evaluates fine-grained visual understanding across six datasets, including instance-level recognition of retail products.
We also report rank correlations between overall scores on \textsc{Pure} and those on each external benchmark across the same ten models. Model-wise scores are provided in Appendix Table~\ref{tab:external-model-results}. %

The resulting correlations are strong for both \textbf{M4Bench} ($\boldsymbol{\rho=0.855}$) and \textbf{FGVQA} ($\boldsymbol{\rho=0.842}$). These correlations suggest that \textsc{Pure} captures capability differences that are also reflected in independently constructed benchmarks of e-commerce image comparison and cross-domain fine-grained recognition.

\section{Conclusion}
\label{sec:conclusion}

We introduced \textsc{Pure}, a diagnostic benchmark for multi-granularity multimodal identity matching, instantiated with real e-commerce records. Its 14 tasks assess Perception, Understanding, Reasoning, and Judgment at the base and variant levels. Across ten MLLMs, attribute understanding is a relative strength, whereas visual difference localization and variant-level rule judgment remain weaknesses. Modality ablations reveal task-specific evidence dependence. In direct matching, most models exhibit different recall patterns at the base and variant levels. %
\textsc{Pure} provides a shared benchmark for comparing model capabilities beyond aggregate matching accuracy, guiding progress toward more reliable multimodal identity matching. %

\section*{Limitations}
\textsc{Pure} evaluates explicit responses from generative MLLMs and therefore does not directly characterize embedding-based retrieval or end-to-end matching systems. The source pool is drawn from difficult cases shared by particular embedding-based and generative matchers, which may inherit their coverage biases. Finally, because the benchmark is instantiated with records from a specific e-commerce setting, generalization across platforms, languages, product categories, and non-commerce identity domains remains to be established.

\section*{Ethics Statement}
  \textsc{Pure} is constructed from real-world e-commerce record pairs containing listing images, titles, and attributes of the selected variant, and is intended for research on multimodal identity matching. Candidate questions were screened using multiple models and verified against the source evidence and task criteria by two experienced e-commerce practitioners. The benchmark may reflect biases in its source systems, category coverage, and listing presentation, and should not be treated as representative of all products or markets.

\bibliography{references}

@inproceedings{m4bench2025,
  title     = {{M4Bench}: A Benchmark of Multi-domain Multi-granularity Multi-image Understanding for Multi-modal Large Language Models},
  author    = {Ye, Xiaojun and Liang, Guanbao and Wang, Chun and Li, Liangcheng and Ke, Pengfei and Wang, Rui and Jia, Bingxin and Huang, Gang and Sun, Qiao and Zhou, Sheng},
  booktitle = {Proceedings of the Thirty-Fourth International Joint Conference on Artificial Intelligence},
  pages     = {6848--6856},
  year      = {2025},
  doi       = {10.24963/ijcai.2025/762},
  url       = {https://www.ijcai.org/proceedings/2025/762}
}

@inproceedings{marsili2026twin,
  title     = {Same or Not? Enhancing Visual Perception in Vision-Language Models},
  author    = {Marsili, Damiano and Mehta, Aditya and Lin, Ryan Y. and Gkioxari, Georgia},
  booktitle = {Proceedings of the IEEE/CVF Conference on Computer Vision and Pattern Recognition},
  pages     = {17303--17315},
  year      = {2026},
  url       = {https://openaccess.thecvf.com/content/CVPR2026/html/Marsili_Same_or_Not_Enhancing_Visual_Perception_in_Vision-Language_Models_CVPR_2026_paper.html}
}

@article{rair2025,
  title={{RAIR}: A Rule-Aware Benchmark Uniting Challenging Long-Tail and Visual Salience Subset for E-commerce Relevance Assessment},
  author={Lu, Chenji and Chen, Zhuo and Zhao, Hui and Wang, Zhenyi and Wang, Pengjie and Yu, Chuan and Xu, Jian},
  journal={arXiv preprint arXiv:2512.24943},
  year={2025},
  url = {https://arxiv.org/abs/2512.24943}
}

@article{oxyecombench2026,
  title   = {{OxyEcomBench}: Benchmarking Multimodal Foundation Models across E-Commerce Ecosystems},
  author  = {Liu, Yong and Liu, Ximan and Yang, Guoqing and Bai, Bing and Xu, Xiaoqiang and Chen, Zhen and Zhang, Ke and Li, Yan},
  journal = {arXiv preprint arXiv:2605.13173},
  year    = {2026},
  url     = {https://arxiv.org/abs/2605.13173}
}

@inproceedings{ecommmmu2025,
  title     = {{EcomMMMU}: Strategic Utilization of Visuals for Robust Multimodal E-commerce Models},
  author    = {Ling, Xinyi and Du, Hanwen and Zhu, Zhihui and Ning, Xia},
  booktitle = {Proceedings of the 14th International Joint Conference on Natural Language Processing and the 4th Conference of the Asia-Pacific Chapter of the Association for Computational Linguistics},
  pages     = {769--790},
  year      = {2025},
  doi       = {10.18653/v1/2025.ijcnlp-long.43},
  url       = {https://aclanthology.org/2025.ijcnlp-long.43/}
}

@inproceedings{naturalbench2024,
  title     = {{NaturalBench}: Evaluating Vision-Language Models on Natural Adversarial Samples},
  author    = {Li, Baiqi and Lin, Zhiqiu and Peng, Wenxuan and Nyandwi, Jean de Dieu and Jiang, Daniel and Ma, Zixian and Khanuja, Simran and Krishna, Ranjay and Neubig, Graham and Ramanan, Deva},
  booktitle = {Advances in Neural Information Processing Systems},
  volume    = {37},
  pages = {17044--17068},       
  year      = {2024},
  doi       = {10.52202/079017-0542},
  url       = {https://proceedings.neurips.cc/paper_files/paper/2024/hash/1e69ff56d0ebff0752ff29caaddc25dd-Abstract-Datasets_and_Benchmarks_Track.html}
}

@inproceedings{taosr12025,
  title     = {{TaoSR1}: The Thinking Model for E-commerce Relevance Search},
  author    = {Dong, Chenhe and Yao, Shaowei and Jiao, Pengkun and Yang, Jianhui and Jin, Yiming and Huang, Zerui and Zhou, Xiaojiang and Ou, Dan and Tang, Haihong and Zheng, Bo},
  pages = {7174--7184}, 
  booktitle = {Proceedings of the 32nd ACM SIGKDD Conference on Knowledge Discovery and Data Mining V.2},
  year      = {2026},
  doi       = {10.1145/3770855.3818489},
  url       = {https://arxiv.org/abs/2508.12365}
}

@inproceedings{vlmsubtlebench2026,
  title     = {{VLM-SubtleBench}: How Far Are {VLM}s from Human-Level Subtle Comparative Reasoning?},
  author    = {Kim, Minkyu and Lee, Sangheon and Park, Dongmin},
  booktitle = {The Fourteenth International Conference on Learning Representations},
  year      = {2026},
  pages = {106840--106868},
  url       = {https://openreview.net/forum?id=pBTXsu1i77}
}

@misc{anthropic2026claudeopus5,
  author = {{Anthropic}},
  title  = {Introducing {Claude Opus 5}},
  year   = {2026},
  month  = jul,
  url    = {https://www.anthropic.com/news/claude-opus-5}
}

@misc{anthropic2026claudeopus46,
  author = {{Anthropic}},
  title  = {{Introducing Claude Opus 4.6}},
  year   = {2026},
  month  = feb,
  url    = {https://www.anthropic.com/news/claude-opus-4-6}
}

@misc{googledeepmind2026gemini38flash,
  author = {{Google DeepMind}},
  title  = {{Gemini 3.8 Flash Model Card}},
  year   = {2026},
  month  = sep,
  url    = {https://storage.googleapis.com/deepmind-media/Model-Cards/Gemini-3-8-Flash-Model-Card.pdf}
}

@misc{googledeepmind2026gemini31pro,
  author = {{Google DeepMind}},
  title  = {{Gemini 3.1 Pro Model Card}},
  year   = {2026},
  month  = feb,
  url    = {https://deepmind.google/models/model-cards/gemini-3-1-pro/}
}

@misc{openai2026gpt56,
  author = {{OpenAI}},
  title  = {{GPT-5.6 System Card}},
  year   = {2026},
  month  = jul,
  url    = {https://deploymentsafety.openai.com/gpt-5-6}
}

@misc{openai2026gpt54,
  author = {{OpenAI}},
  title  = {Introducing {GPT-5.4}},
  year   = {2026},
  month  = mar,
  url    = {https://openai.com/index/introducing-gpt-5-4/}
}

@misc{qwen2026qwen38,
  author = {{Qwen Team}},
  title  = {{Qwen3.8-Max}: A New Bar for Coding and Cowork},
  year   = {2026},
  month  = aug,
  url    = {https://qwen.ai/blog?id=qwen3.8}
}

@misc{qwen2026qwen35,
  author = {{Qwen Team}},
  title  = {{Qwen3.5}: Towards Native Multimodal Agents},
  year   = {2026},
  month  = feb,
  url    = {https://qwen.ai/blog?id=qwen3.5}
}

@misc{deepseekai2026deepseekv41flash,
  title         = {DeepSeek-V4.1-Flash: Pushing the Limits of KV Cache Compression},
  author        = {{DeepSeek-AI}},
  year          = {2026},
  eprint        = {2609.19969},
  archivePrefix = {arXiv},
  primaryClass  = {cs.CL},
  url           = {https://arxiv.org/abs/2609.19969}
}

@misc{kimiteam2026kimik3,
  author        = {{Kimi Team}},
  title         = {{Kimi K3}: Open Frontier Intelligence},
  year          = {2026},
  eprint        = {2607.24653},
  archivePrefix = {arXiv},
  primaryClass  = {cs.CL},
  doi           = {10.48550/arXiv.2607.24653},
  url           = {https://arxiv.org/abs/2607.24653}
}

@inproceedings{zhan2021product1m,
  title     = {{Product1M}: Towards Weakly Supervised Instance-Level Product Retrieval via Cross-Modal Pretraining},
  author    = {Zhan, Xunlin and Wu, Yangxin and Dong, Xiao and Wei, Yunchao and Lu, Minlong and Zhang, Yichi and Xu, Hang and Liang, Xiaodan},
  booktitle = {Proceedings of the IEEE/CVF International Conference on Computer Vision},
  pages     = {11782--11791},
  year      = {2021},
  url       = {https://openaccess.thecvf.com/content/ICCV2021/html/Zhan_Product1M_Towards_Weakly_Supervised_Instance-Level_Product_Retrieval_via_Cross-Modal_Pretraining_ICCV_2021_paper.html}
}

@inproceedings{dong2022m5product,
  title     = {{M5Product}: Self-Harmonized Contrastive Learning for E-Commercial Multi-Modal Pretraining},
  author    = {Dong, Xiao and Zhan, Xunlin and Wu, Yangxin and Wei, Yunchao and Kampffmeyer, Michael C. and Wei, Xiaoyong and Lu, Minlong and Wang, Yaowei and Liang, Xiaodan},
  booktitle = {Proceedings of the IEEE/CVF Conference on Computer Vision and Pattern Recognition},
  pages     = {21252--21262},
  year      = {2022},
  url       = {https://openaccess.thecvf.com/content/CVPR2022/html/Dong_M5Product_Self-Harmonized_Contrastive_Learning_for_E-Commercial_Multi-Modal_Pretraining_CVPR_2022_paper.html}
}

@inproceedings{fu2024blink,
  title     = {{BLINK}: Multimodal Large Language Models Can See but Not Perceive},
  author    = {Fu, Xingyu and Hu, Yushi and Li, Bangzheng and Feng, Yu and Wang, Haoyu and Lin, Xudong and Roth, Dan and Smith, Noah A. and Ma, Wei-Chiu and Krishna, Ranjay},
  booktitle = {European Conference on Computer Vision},
  pages     = {148--166},
  year      = {2024},
  publisher = {Springer},
  url       = {https://www.ecva.net/papers/eccv_2024/papers_ECCV/html/3356_ECCV_2024_paper.php}
}

@inproceedings{liu2024mibench,
  title     = {{MIBench}: Evaluating Multimodal Large Language Models over Multiple Images},
  author    = {Liu, Haowei and Zhang, Xi and Xu, Haiyang and Shi, Yaya and Jiang, Chaoya and Yan, Ming and Zhang, Ji and Huang, Fei and Yuan, Chunfeng and Li, Bing and Hu, Weiming},
  booktitle = {Proceedings of the 2024 Conference on Empirical Methods in Natural Language Processing},
  pages     = {22417--22428},
  year      = {2024},
  doi       = {10.18653/v1/2024.emnlp-main.1250},
  url       = {https://aclanthology.org/2024.emnlp-main.1250/}
}

@inproceedings{kil2024mllmcompbench,
  title     = {{MLLM-CompBench}: A Comparative Reasoning Benchmark for Multimodal {LLM}s},
  author    = {Kil, Jihyung and Mai, Zheda and Lee, Justin and Chowdhury, Arpita and Wang, Zihe and Cheng, Kerrie and Wang, Lemeng and Liu, Ye and Chao, Wei-Lun},
  booktitle = {Advances in Neural Information Processing Systems},
  volume    = {37},
  pages={28798--28827},
  year      = {2024},
  url       = {https://proceedings.neurips.cc/paper_files/paper/2024/hash/32923dff09f75cf1974c145764a523e2-Abstract-Datasets_and_Benchmarks_Track.html}
}

@inproceedings{jin2024shoppingmmlu,
  title         = {Shopping {MMLU}: A Massive Multi-Task Online Shopping Benchmark for Large Language Models},
  author        = {Jin, Yilun and Li, Zheng and Zhang, Chenwei and Cao, Tianyu and Gao, Yifan and Jayarao, Pratik and Li, Mao and Liu, Xin and Sarkhel, Ritesh and Tang, Xianfeng and Wang, Haodong and Wang, Zhengyang and Xu, Wenju and Yang, Jingfeng and Yin, Qingyu and Li, Xian and Nigam, Priyanka and Xu, Yi and Chen, Kai and Yang, Qiang and Jiang, Meng and Yin, Bing},
  booktitle = {Advances in Neural Information Processing Systems},
  doi = {10.52202/079017-0574},
  volume={37},
  pages = {18062--18089},
  year          = {2024},
  url           = {https://proceedings.neurips.cc/paper_files/paper/2024/file/2049d75dd13db049897562bcf7d59da8-Paper-Datasets_and_Benchmarks_Track.pdf}
}

@article{zhang2026fgpr,
  title   = {{FGPR}: A Large-Scale Dataset and Benchmark for Fine-Grained Product Retrieval},
  author  = {Zhang, Ruisong and Hu, Jie and Li, Zhong-Zhi and Wang, Chuang and Liu, Cheng-Lin},
  journal = {Pattern Recognition},
  volume  = {172},
  pages   = {112523},
  year    = {2026},
  doi     = {10.1016/j.patcog.2025.112523},
  url     = {https://www.sciencedirect.com/science/article/pii/S0031320325011860}
}

@inproceedings{deng2025words,
  author    = {Deng, Ailin and Cao, Tri and Chen, Zhirui and Hooi, Bryan},
  title     = {Words or Vision: Do Vision-Language Models Have Blind Faith in Text?},
  booktitle = {Proceedings of the IEEE/CVF Conference on Computer Vision and Pattern Recognition},
  pages     = {3867--3876},
  year      = {2025},
  url       = {https://openaccess.thecvf.com/content/CVPR2025/html/Deng_Words_or_Vision_Do_Vision-Language_Models_Have_Blind_Faith_in_CVPR_2025_paper.html}
}

@article{spearman1904proof,
 URL = {http://www.jstor.org/stable/1412159},
 author = {C. Spearman},
 journal = {The American Journal of Psychology},
 number = {1},
 pages = {72--101},
 title = {The Proof and Measurement of Association between Two Things},
 volume = {15},
 year = {1904}
}

@article{smith2025facevoice,
  title   = {Face and voice identity matching accuracy is not improved by multimodal identity information},
  author  = {Smith, Harriet M. J. and Ritchie, Kay L. and Baguley, Thom S. and Lavan, Nadine},
  journal = {British Journal of Psychology},
  volume  = {116},
  number  = {2},
  pages   = {367--385},
  year    = {2025},
  doi     = {10.1111/bjop.12757},
  url     = {https://doi.org/10.1111/bjop.12757}
}

@article{henderson2025dynamic,
  title   = {Dynamic categorization rules alter representations in human visual cortex},
  author  = {Henderson, Margaret M. and Serences, John T. and Rungratsameetaweemana, Nuttida},
  journal = {Nature Communications},
  volume  = {16},
  number  = {1},
  pages   = {3459},
  year    = {2025},
  doi     = {10.1038/s41467-025-58707-4},
  url     = {https://www.nature.com/articles/s41467-025-58707-4}
}

@inproceedings{vaze2023genecis,
  title     = {{GeneCIS}: A Benchmark for General Conditional Image Similarity},
  author    = {Vaze, Sagar and Carion, Nicolas and Misra, Ishan},
  booktitle = {Proceedings of the IEEE/CVF Conference on Computer Vision and Pattern Recognition},
  pages     = {6862--6872},
  year      = {2023},
  url       = {https://openaccess.thecvf.com/content/CVPR2023/html/Vaze_GeneCIS_A_Benchmark_for_General_Conditional_Image_Similarity_CVPR_2023_paper.html}
}

@article{chen2026pailitaovl,
  title   = {{Pailitao-VL}: Unified Embedding and Reranker for Real-Time Multi-Modal Industrial Search},
  author  = {Chen, Lei and Ju, Chen and Chen, Xu and Wang, Zhicheng and Jiao, Yuheng and Zhan, Hongfeng and Li, Zhaoyang and Xu, Shihao and Zhao, Zhixiang and Jia, Tong and Li, Lin and Gao, Yuan and Song, Jun and Lan, Jinsong and Zhu, Xiaoyong and Zheng, Bo},
  journal = {arXiv preprint arXiv:2602.13704},
  year    = {2026},
  doi     = {10.48550/arXiv.2602.13704},
  url     = {https://arxiv.org/abs/2602.13704}
}

@article{industrybench2026,
  title   = {{IndustryBench}: Probing the Industrial Knowledge Boundaries of {LLMs}},
  author  = {Bai, Songlin and Wang, Xintong and Yu, Linlin and Chen, Bin and Xu, Zhiang and Sheng, Yuyang and Zan, Changtong and Zhu, Xiaofeng and Zhang, Yizhe and Li, Jiru and Guo, Mingze and Zou, Ling and Li, Yalong and Huo, Chengfu and Ding, Liang},
  journal = {arXiv preprint arXiv:2605.10267},
  year    = {2026},
  doi     = {10.48550/arXiv.2605.10267},
  url     = {https://arxiv.org/abs/2605.10267}
}

@article{industrybenchmipu2026,
  title   = {{IndustryBench-MIPU}: Benchmarking Multi-Image Attribute Value Extraction for Industrial Products},
  author  = {Qi, Haonan and Cao, Jin and Zhang, Yongqi and Wang, Xintong and Tang, Weidong and Chen, Bin and Huo, Chengfu and Pan, Haojun and You, Hengyu and Li, Jing and Wang, Yingde and Ding, Liang},
  journal = {arXiv preprint arXiv:2606.14383},
  year    = {2026},
  doi     = {10.48550/arXiv.2606.14383},
  url     = {https://arxiv.org/abs/2606.14383}
}

\appendix

\section{Task Definitions} %
\label{app:task-definitions}

\begin{figure*}[t]
  \centering
  \includegraphics[width=\textwidth]{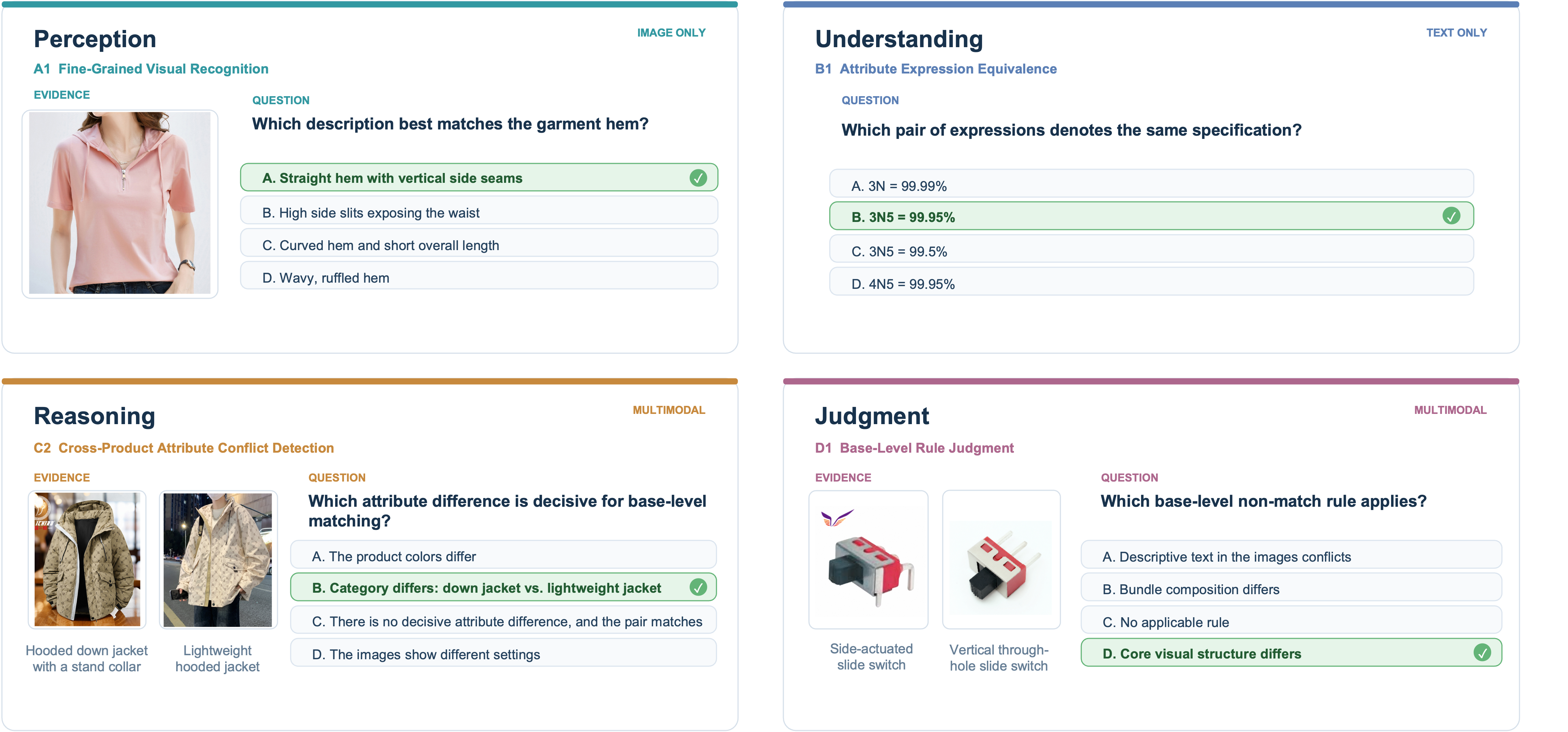} %
  \caption{\textbf{Representative questions illustrating the four diagnostic dimensions of \textsc{Pure}.} Each panel presents a question and its answer options, with the correct option highlighted in green.}
  \label{fig:representative-questions}
\end{figure*}

Table~\ref{tab:taxonomy} summarizes the complete task taxonomy. Figure~\ref{fig:representative-questions} illustrates representative questions from A1, B1, C2, and D1, covering the four diagnostic dimensions.

\begin{table*}[t]
  \centering
   \caption{\textbf{Task taxonomy of \textsc{Pure}.} \textbf{Gran.} denotes the target identity-matching granularity. \textbf{T}, \textbf{SI}, and \textbf{MI} denote textual evidence, a single image, and multiple images, respectively. Combinations indicate joint evidence. Task identifiers are shown in parentheses and encode dimension membership.}
  \label{tab:taxonomy}
  \footnotesize
  \setlength{\tabcolsep}{4pt}
  \renewcommand{\arraystretch}{1.12}
  \begin{tabular}{@{}p{0.13\textwidth}p{0.07\textwidth}p{0.245\textwidth}p{0.075\textwidth}p{0.38\textwidth}@{}}
    \hline
    \textbf{Dimension} & \textbf{Gran.} & \textbf{Task} & \textbf{Modality}& \textbf{Primary diagnostic focus} \\
    \hline
    \multirow[c]{3}{*}{\textbf{Perception}} & Base & Fine-Grained Visual Recognition \textit{(A1)} & SI & Fine-grained appearance and structure. \\
      & Base & Product Image OCR \textit{(A2)} & SI & Text printed on products, packaging, labels, or nameplates. \\
      & Base & Image--Text Evidence Extraction \textit{(A3)} & T+SI & Attributes supported by combined visual and textual evidence. \\
    \hline
    \multirow[c]{5}{*}{\textbf{Understanding}} & Base & Attribute Expression Equivalence \textit{(B1)} & T & Semantic equivalence of attribute expressions. \\
      & Base & Primary Product and Bundle Relations \textit{(B2)} & T+SI & Primary-product, accessory, and bundle roles. \\
      & Base & Product Title Attribute Interpretation \textit{(B3)} & T & Product terms, model expressions, and title attributes. \\
      & Variant & Textual Core Attribute Extraction \textit{(B4)} & T & Core attributes of the selected variant under title--variant ambiguity or conflict. \\
      & Variant & Variant-Level Attribute Equivalence \textit{(B5)} & T & Attribute equivalence between two selected variants.\\
    \hline
    \multirow[c]{4}{*}{\textbf{Reasoning}} & Base & Visual Difference Localization \textit{(C1)} & MI & Presence and location of identity-relevant visual differences. \\
      & Base & Cross-Product Attribute Conflict Detection \textit{(C2)} & T+MI & Attribute conflicts across two product listings. \\
      & Base & Contrastive Attribute Grounding \textit{(C3)}& T+SI & Verification over a $2 \times 2$ product--attribute group.\\
      & Variant & Variant-Level Visual Equivalence \textit{(C4)}& MI & Variant-level identity under presentation variation and genuine visual differences. \\
    \hline
    \multirow[c]{2}{*}{\textbf{Judgment}} & Base & Base-Level Rule Judgment \textit{(D1)} & T+MI & Applicable base-level non-match rule.\\
      & Variant & Variant-Level Rule Judgment \textit{(D2)}& T+MI & Applicable variant-level non-match rule. \\
    \hline
  \end{tabular}
  
\end{table*}

\section{Formal Metric Definitions}
\label{app:metric-definitions}

For each task $t$, the task accuracy is computed as
\begin{equation}
  \mathrm{Acc}_t
  = \frac{1}{N_t}\sum_{i=1}^{N_t}
    \mathbb{I}[\hat{y}_{t,i}=y_{t,i}],
  \label{eq:accuracy}
\end{equation}
where \(N_t\) is the number of questions in task \(t\), and \(\hat{y}_{t,i}\) and \(y_{t,i}\) denote the predicted and reference answers, respectively. For the multiple-selection tasks A2 and A3, answers are deterministically parsed and normalized before comparison. A response is correct only when the predicted answer exactly matches the reference answer. %

C3 is evaluated at the group level. Each group contains four binary questions and is counted as correct only when all four responses are correct~\citep{naturalbench2024}:
\begin{equation}
  \mathrm{G\text{-}Acc}
  = \frac{1}{G}\sum_{g=1}^{G}
    \prod_{j=1}^{4}
    \mathbb{I}[\hat{y}_{g,j}=y_{g,j}],
  \label{eq:gacc}
\end{equation}
where \(G\) is the number of C3 groups.

Let \(s_t\) denote the score for task \(t\). We set \(s_t=\mathrm{Acc}_t\) for the 13 tasks other than C3, and \(s_{\mathrm{C3}}=\mathrm{G\text{-}Acc}\) for C3. %
Let \(\mathcal{T}\) denote the full set of 14 tasks and \(\mathcal{T}_d\) the tasks associated with dimension \(d\). We report a dimension score for each dimension and an overall score, both computed as equal-weight averages over their constituent task scores:
\begin{align}
  S_d
  &= \frac{1}{|\mathcal{T}_d|}
     \sum_{t\in\mathcal{T}_d} s_t,
  \label{eq:capability-score}\\
  S_{\mathrm{overall}}
  &= \frac{1}{|\mathcal{T}|}
     \sum_{t\in\mathcal{T}} s_t.
  \label{eq:overall-score}
\end{align}

\section{Additional Results}
\label{app:additional-results}

\subsection{Effect of Thinking}

\label{app:thinking}
\label{sec:thinking-ablation}

We compare Qwen3.8-Max, DeepSeek-V4.1-Flash, and the Qwen3.5-9B, 27B, 35B-A3B, and 397B-A17B variants with thinking disabled and enabled. Within each model, the two modes use the same 1,457 questions, product evidence, task prompts, and answer schema. Task scores and their aggregation follow Section~\ref{sec:metrics}.

\begin{table*}[t]
  \centering
  \definecolor{thinkingcandidategain}{HTML}{EAF5EC}
  \definecolor{thinkingcandidateloss}{HTML}{FBECEC}
  \definecolor{thinkingcandidategaintext}{HTML}{2F7551}
  \definecolor{thinkingcandidatelosstext}{HTML}{A64444}
  \caption{\textbf{Task-level performance with and without thinking on \textsc{Pure} (\%).} Non-thinking scores reproduce Table~\ref{tab:main-consolidated-grouped-candidate}. Green and red cells indicate higher and lower scores than the same model without thinking. Unchanged scores are unshaded. Overall changes, computed before rounding, are shown in parentheses in percentage points. C3 uses group accuracy.}
  \label{tab:thinking-task-results-candidate}
  \scriptsize
  \setlength{\tabcolsep}{3.0pt} %
  \renewcommand{\arraystretch}{1.3} %
  \begin{tabular}{@{}lc*{14}{r}r@{\,}l@{}}
    \toprule
    \multirow{2}{*}{\textbf{Model}}
      & \multirow{2}{*}{\textbf{Thinking}}
      & \multicolumn{3}{c}{\textbf{Perception}}
      & \multicolumn{5}{c}{\textbf{Understanding}}
      & \multicolumn{4}{c}{\textbf{Reasoning}}
      & \multicolumn{2}{c}{\textbf{Judgment}}
      & \multicolumn{2}{c@{}}{\multirow{2}{*}{\textbf{Overall ($\Delta$)}}} \\
    \cmidrule(lr){3-5}\cmidrule(lr){6-10}
    \cmidrule(lr){11-14}\cmidrule(lr){15-16}
      & & A1 & A2 & A3 & B1 & B2 & B3 & B4 & B5
      & C1 & C2 & C3 & C4 & D1 & D2 & & \\
    \midrule
    \multirow[c]{2}{*}{\textbf{Qwen3.8-Max}}
      & w/o & 87.10 & 64.23 & 71.15 & 65.76 & 93.14 & 88.00 & 96.04 & 97.03 & 47.89 & 63.41 & 76.67 & 96.15 & 73.38 & 57.60 & 76.97 & \\
      & w/  & \cellcolor{thinkingcandidateloss}81.72 & \cellcolor{thinkingcandidategain}81.30 & \cellcolor{thinkingcandidateloss}68.27 & \cellcolor{thinkingcandidategain}80.98 & \cellcolor{thinkingcandidategain}98.04 & \cellcolor{thinkingcandidateloss}85.33 & 96.04 & \cellcolor{thinkingcandidategain}99.01 & \cellcolor{thinkingcandidategain}70.42 & \cellcolor{thinkingcandidategain}70.73 & \cellcolor{thinkingcandidategain}90.00 & \cellcolor{thinkingcandidateloss}92.31 & \cellcolor{thinkingcandidategain}76.26 & 57.60 & \textbf{82.00} & {\textcolor{thinkingcandidategaintext}{(+5.03)}} \\
    \midrule
    \multirow[c]{2}{*}{\textbf{DeepSeek-V4.1-Flash}}
      & w/o & 70.97 & 47.97 & 72.12 & 61.96 & 92.16 & 77.33 & 96.04 & 95.05 & 32.39 & 80.49 & 60.00 & 84.62 & 67.63 & 36.00 & 69.62 & \\
      & w/  & \cellcolor{thinkingcandidateloss}66.67 & \cellcolor{thinkingcandidategain}77.24 & \cellcolor{thinkingcandidategain}79.81 & \cellcolor{thinkingcandidategain}84.24 & \cellcolor{thinkingcandidategain}93.14 & \cellcolor{thinkingcandidategain}90.67 & \cellcolor{thinkingcandidateloss}94.06 & \cellcolor{thinkingcandidategain}99.01 & \cellcolor{thinkingcandidategain}63.38 & \cellcolor{thinkingcandidateloss}70.73 & \cellcolor{thinkingcandidategain}70.00 & 84.62 & \cellcolor{thinkingcandidateloss}58.99 & \cellcolor{thinkingcandidategain}38.40 & \textbf{76.50} & {\textcolor{thinkingcandidategaintext}{(+6.87)}} \\
    \midrule
    \multirow[c]{2}{*}{\textbf{Qwen3.5-397B-A17B}}
      & w/o & 69.89 & 48.78 & 43.27 & 46.74 & 85.29 & 61.33 & 97.03 & 97.03 & 39.44 & 26.83 & 50.00 & 80.77 & 45.32 & 37.60 & 59.24 & \\
      & w/  & \cellcolor{thinkingcandidateloss}64.52 & \cellcolor{thinkingcandidategain}74.80 & \cellcolor{thinkingcandidategain}55.77 & \cellcolor{thinkingcandidategain}72.28 & \cellcolor{thinkingcandidategain}89.22 & \cellcolor{thinkingcandidategain}76.00 & \cellcolor{thinkingcandidateloss}95.05 & \cellcolor{thinkingcandidategain}99.01 & \cellcolor{thinkingcandidategain}53.52 & \cellcolor{thinkingcandidategain}82.93 & \cellcolor{thinkingcandidategain}63.33 & \cellcolor{thinkingcandidategain}82.05 & \cellcolor{thinkingcandidategain}64.75 & \cellcolor{thinkingcandidategain}49.60 & \textbf{73.06} & {\textcolor{thinkingcandidategaintext}{(+13.82)}} \\
    \midrule
    \multirow[c]{2}{*}{\textbf{Qwen3.5-27B}}
      & w/o & 51.61 & 52.03 & 55.77 & 35.33 & 81.37 & 53.33 & 98.02 & 97.03 & 25.35 & 60.98 & 33.33 & 83.33 & 51.08 & 36.80 & 58.24 & \\
      & w/  & 51.61 & \cellcolor{thinkingcandidategain}73.17 & \cellcolor{thinkingcandidateloss}52.88 & \cellcolor{thinkingcandidategain}55.98 & \cellcolor{thinkingcandidategain}85.29 & \cellcolor{thinkingcandidategain}66.67 & \cellcolor{thinkingcandidateloss}95.05 & \cellcolor{thinkingcandidategain}100.00 & \cellcolor{thinkingcandidategain}46.48 & \cellcolor{thinkingcandidategain}70.73 & \cellcolor{thinkingcandidategain}66.67 & \cellcolor{thinkingcandidategain}89.74 & \cellcolor{thinkingcandidategain}69.06 & \cellcolor{thinkingcandidategain}49.60 & \textbf{69.50} & {\textcolor{thinkingcandidategaintext}{(+11.26)}} \\
    \midrule
    \multirow[c]{2}{*}{\textbf{Qwen3.5-35B-A3B}}
      & w/o & 36.56 & 42.28 & 49.04 & 27.72 & 79.41 & 38.67 & 95.05 & 84.16 & 9.86 & 43.90 & 50.00 & 83.33 & 38.85 & 35.20 & 51.00 & \\
      & w/  & \cellcolor{thinkingcandidategain}54.84 & \cellcolor{thinkingcandidategain}73.98 & \cellcolor{thinkingcandidategain}51.92 & \cellcolor{thinkingcandidategain}57.07 & \cellcolor{thinkingcandidategain}83.33 & \cellcolor{thinkingcandidategain}57.33 & \cellcolor{thinkingcandidateloss}93.07 & \cellcolor{thinkingcandidategain}100.00 & \cellcolor{thinkingcandidategain}39.44 & \cellcolor{thinkingcandidategain}48.78 & \cellcolor{thinkingcandidategain}53.33 & \cellcolor{thinkingcandidategain}89.74 & \cellcolor{thinkingcandidategain}54.68 & \cellcolor{thinkingcandidategain}40.80 & \textbf{64.17} & {\textcolor{thinkingcandidategaintext}{(+13.16)}} \\
    \midrule
    \multirow[c]{2}{*}{\textbf{Qwen3.5-9B}}
      & w/o & 51.61 & 29.27 & 37.50 & 31.52 & 68.63 & 18.67 & 89.11 & 89.11 & 22.54 & 36.59 & 26.67 & 76.92 & 45.32 & 36.00 & 47.10 & \\
      & w/  & 51.61 & \cellcolor{thinkingcandidategain}64.23 & \cellcolor{thinkingcandidategain}50.00 & \cellcolor{thinkingcandidategain}42.93 & \cellcolor{thinkingcandidategain}72.55 & \cellcolor{thinkingcandidategain}28.00 & \cellcolor{thinkingcandidategain}95.05 & \cellcolor{thinkingcandidategain}99.01 & \cellcolor{thinkingcandidategain}33.80 & \cellcolor{thinkingcandidategain}48.78 & \cellcolor{thinkingcandidategain}43.33 & \cellcolor{thinkingcandidateloss}73.08 & \cellcolor{thinkingcandidategain}53.96 & \cellcolor{thinkingcandidateloss}30.40 & \textbf{56.20} & {\textcolor{thinkingcandidategaintext}{(+9.09)}} \\
    \bottomrule
  \end{tabular}
\end{table*}

Table~\ref{tab:thinking-task-results-candidate} shows overall gains of 5.0--13.8 percentage points across the six models. All models improve on C1 and C3, but improvements do not extend to every task. We analyzed DeepSeek-V4.1-Flash's incorrect responses on C2 and D1 with thinking enabled. This analysis suggests that the model sometimes treats missing attribute information or differences unrelated to identity as evidence of a conflict.

\begin{table}[t] %
  \centering
    \caption{\textbf{Mean accuracy (\%) and accuracy differences (percentage points), averaged across all ten MLLMs.} $N$ is the number of unique questions per model. Positive $\Delta_m$ indicates a decrease in accuracy after evidence removal, and negative $\Delta_m$ indicates an increase.}
  \small
  \setlength{\tabcolsep}{2pt}%
   \begin{tabular*}{\columnwidth}{@{\extracolsep{\fill}}lrrrrrr@{}}
    \toprule
    \textbf{Task} & \textbf{$N$} & \textbf{Full} & \textbf{w/o Images} & \textbf{w/o Titles} & \textbf{$\Delta_{\mathrm{image}}$} & \textbf{$\Delta_{\mathrm{title}}$} \\
    \midrule
    A3 & 104 & 63.6 & 48.8 & 40.6 & +14.7 & +23.0 \\
    C2 & 41 & 58.5 & 49.8 & 61.2 & +8.8 & $-2.7$ \\
    D1 & 139 & 61.2 & 43.7 & 69.4 & +17.5 & $-8.2$ \\
    \bottomrule
  \end{tabular*}
  \label{tab:rq2-aggregate}
\end{table} %

\subsection{Modality-Ablation Results}
\label{app:rq2-details}

Table~\ref{tab:rq2-aggregate} reports the mean accuracy across all ten MLLMs. The \emph{w/o Images} condition removes all listing images, whereas the \emph{w/o Titles} condition removes all listing titles text. Task instructions and answer options remain fixed across conditions. We use $\Delta_m=\mathrm{Acc}_{\mathrm{Full}}-\mathrm{Acc}_{\mathrm{Ablated}}$ as defined in Section~\ref{sec:rq2}, with a positive value indicating a decrease in accuracy after evidence removal.

Table~\ref{tab:d1-stratified} groups D1 questions by the correct answer. "No applicable rule" corresponds to matching pairs, whereas the other three answer categories correspond to non-matching pairs. Removing listing titles increases mean accuracy from 48.3\% to 71.7\% when the correct answer is "no applicable rule" but decreases it when the correct answer specifies a non-match rule. %

\begin{table}[t] %
  \centering
  \caption{\textbf{D1 accuracy grouped by the correct answer, averaged across all ten models (\%).} "No applicable rule" corresponds to matching pairs, whereas the other three answer categories correspond to non-matching pairs.}
  \footnotesize
  \setlength{\tabcolsep}{4pt} %
    \begin{tabular*}{\columnwidth}{@{}l@{\hspace{8pt}}r@{\extracolsep{\fill}}rr@{}}
    \toprule
   \textbf{Correct answer} & \textbf{Full} & \begin{tabular}[c]{@{}c@{}}\textbf{w/o}\\\textbf{Images}\end{tabular} & \begin{tabular}[c]{@{}c@{}}\textbf{w/o}\\\textbf{Titles}\end{tabular} \\
    \midrule
    No applicable rule & 48.3 & 28.8 & 71.7 \\
    Core visual structure differs & 75.1 & 58.6 & 74.3 \\
    Titles refer to different products & 86.7 & 85.0 & 13.3 \\
    Bundle composition differs & 78.8 & 62.5 & 58.8 \\
    \bottomrule
  \end{tabular*}
  \label{tab:d1-stratified}
\end{table} %

\subsection{Direct Base-Level and Variant-Level Matching Results}
\label{app:direct-matching}

Table~\ref{tab:direct-matching-results} reports the direct-matching results used in RQ3. The evaluation set contains approximately equal numbers of matching and non-matching pairs. Balanced accuracy is the mean of match and non-match recall. 

\begin{table*}[t]
  \centering
  \caption{\textbf{Direct base-level and variant-level identity matching on the post-QC frozen subset (\%).} M and NM denote matching and non-matching pairs. BAcc. is balanced accuracy, and $R_{\mathrm{M}}$ and $R_{\mathrm{NM}}$ are class-specific recalls.}
  
  \small
  \setlength{\tabcolsep}{4.5pt}
  \begin{tabular}{@{}l rrr rrr@{}}
    \toprule
    & \multicolumn{3}{c}{\textbf{Base (165 M / 165 NM)}} & \multicolumn{3}{c}{\textbf{Variant (103 M / 108 NM)}} \\
    \cmidrule(lr){2-4} \cmidrule(lr){5-7}
    \textbf{Model} & \textbf{BAcc.} & \textbf{$R_{\mathrm{M}}$} & \textbf{$R_{\mathrm{NM}}$} & \textbf{BAcc.} & \textbf{$R_{\mathrm{M}}$} & \textbf{$R_{\mathrm{NM}}$} \\
    \midrule
    Gemini 3.8 Flash      & 80.91 & 90.91 & 70.91 & 74.91 & 56.31 & 93.52 \\
    GPT-5.6 Sol           & 68.48 & 67.27 & 69.70 & 72.49 & 51.46 & 93.52 \\
    Qwen3.8-Max           & 68.79 & 81.82 & 55.76 & 73.20 & 62.14 & 84.26 \\
    Claude Opus 5         & 72.73 & 73.94 & 71.52 & 66.72 & 62.14 & 71.30 \\
    \midrule
    Kimi K3               & 63.03 & 72.12 & 53.94 & 60.59 & 57.28 & 63.89 \\
    DeepSeek-V4.1-Flash   & 58.48 & 57.58 & 59.39 & 60.71 & 42.72 & 78.70 \\
    Qwen3.5-397B-A17B     & 50.30 & 54.55 & 46.06 & 55.73 & 47.57 & 63.89 \\
    Qwen3.5-27B           & 60.61 & 77.58 & 43.64 & 53.02 & 50.49 & 55.56 \\
    Qwen3.5-35B-A3B       & 50.61 & 70.30 & 30.91 & 48.09 & 57.28 & 38.89 \\
    Qwen3.5-9B            & 50.00 & 63.64 & 36.36 & 45.52 & 46.60 & 44.44 \\
    \midrule
    \textbf{Mean}         & 62.39 & 70.97 & 53.82 & 61.10 & 53.40 & 68.80 \\
    \bottomrule
  \end{tabular}
  \label{tab:direct-matching-results}
\end{table*}

\subsection{External-Benchmark Model-Level Results}
\label{app:external-results}

Table~\ref{tab:external-model-results} reports the model-wise scores underlying the external-benchmark correlations in Section~\ref{sec:rq3}, following VLM-SubtleBench~\citep{vlmsubtlebench2026}.  We use the publicly released datasets of M4Bench and FGVQA, with evaluation settings based on their publicly released scripts.

\begin{table*}[t]
  \centering
  \caption{\textbf{Model-wise overall results used for the external-benchmark analysis (\%).} M4Bench Overall is the unweighted mean of five task accuracies. FGVQA Overall is accuracy over all 12,000 queries from its six equally sized subsets.}
  \small
  \setlength{\tabcolsep}{8pt}
  \begin{tabular}{@{}l r r r@{}}
    \toprule
    \textbf{Model} & \textbf{\textsc{Pure} Overall}& \textbf{M4Bench Overall}& \textbf{FGVQA Overall} \\
    \midrule
    \multicolumn{4}{l}{\textit{Closed-source models}} \\
    Gemini 3.8 Flash & 86.00 & 68.34 & 86.88 \\
    GPT-5.6 Sol & 79.68 & 69.41 & 87.18 \\
    Qwen3.8-Max & 76.97 & 61.94 & 89.99 \\
    Claude Opus 5 & 75.73 & 70.36 & 88.94 \\
    \midrule
    \multicolumn{4}{l}{\textit{Open-source models}} \\
    Kimi K3 & 73.07 & 63.44 & 83.92 \\
    DeepSeek-V4.1-Flash & 69.62 & 53.82 & 81.84 \\
    Qwen3.5-397B-A17B & 59.24 & 56.98 & 81.93 \\
    Qwen3.5-27B & 58.24 & 56.01 & 83.24 \\
    Qwen3.5-35B-A3B & 51.00 & 53.33 & 79.68 \\
    Qwen3.5-9B     & 47.10 & 46.06 & 77.51 \\
    \bottomrule
  \end{tabular}
  \label{tab:external-model-results}
\end{table*}

\section{Human Review Protocol}
\label{app:human-review}

\paragraph{Reviewer background and eligibility.}
The review was conducted by two domain experts, one woman and one man. Both were practitioners at an e-commerce company with multiple years of experience in product understanding and identity-matching work. Their routine work required interpreting listing images, titles, and attributes of the selected variant under operational matching criteria, and they were therefore qualified to verify the evidence and base-level and variant-level decisions represented in \textsc{Pure}. %

\paragraph{Review procedure.}
Each scoring unit that passed multi-model screening was reviewed by both experts against its source record or record pair, the applicable identity criterion, and the assigned task. The reviewers checked (i) whether the supplied evidence was sufficient, (ii) whether the question was self-contained, (iii) whether the options were complete and supported a unique reference answer, and (iv) whether the task assignment and identity granularity were consistent with the source case. Units that failed any criterion were removed. Disagreements were resolved through joint re-examination of the source evidence, and a unit was retained only when both reviewers judged it qualified. C3 was reviewed and retained only as a complete four-question group.

\paragraph{Retention statistics.}
The generation stage produced 3,940 individual questions. Multi-model screening retained 2,819 questions (72\%). Following expert review and final task-wise assembly, the released benchmark contains 1,457 questions and 1,367 scoring units: 1,337 independently scored questions and 30 complete C3 groups, each comprising four questions.

\paragraph{Review instructions.} %
The following checklist summarizes the expert review
procedure.

\begin{promptbox}{Human Review Instructions}
Review each candidate using its source record or record
pair, applicable identity criterion, and assigned task.

\medskip
\textbf{1. Evidence sufficiency.}
Check whether the supplied evidence is sufficient to
answer the question.

\medskip
\textbf{2. Question completeness.}
Check whether the question is self-contained.

\medskip
\textbf{3. Options and reference answer.}
Check whether the options are complete and the reference
answer is uniquely supported by the evidence. For
multiple-answer questions, check the complete set of
correct options.

\medskip
\textbf{4. Task and granularity.}
Check whether the question is consistent with its assigned
task and the identity-matching granularity of the source
case.
For C3, review and retain only complete four-question groups.

\medskip
Reject a candidate that fails any requirement. Resolve
disagreements by jointly re-examining the source evidence.
Retain a candidate only when both reviewers approve it.
\end{promptbox}

\section{Prompts}
\label{app:prompts}
The original prompts are in Chinese. English translations are provided below.

\subsection{Evaluation Prompt}

We present the shared system instruction used for benchmark evaluation.

\begin{promptbox}{System Prompt}
You are an expert in visual understanding of e-commerce
products. Based on the provided information and question,
select the most accurate answer from the given options.

\medskip
For single-choice questions, select only one option letter
(e.g., A). For multiple-answer questions, select all correct
option letters (e.g., ABC).

\medskip
You must strictly follow the JSON format below.
Do not include explanations or Markdown code blocks.

\medskip
\texttt{\{"answer": "Final option letter(s), e.g., A or ABC"\}}
\end{promptbox}

\subsection{Direct Matching Prompt}
\label{app:direct-matching-prompts}

Direct identity matching uses a separate system instruction. The user message specifies the target matching granularity. Option A denotes a
match, and option B denotes a non-match. The terms "Item" and "SKU" in the original prompts correspond to the base and variant levels defined in this paper, respectively.

\begin{promptbox}{Direct Matching Prompt}
You are an evaluator of e-commerce product matching.
Determine whether two products belong to the same Item
or the same SKU.

\medskip
An Item match requires agreement in core category, brand,
product model, structure, and style. Non-essential
differences in seller, image background, packaging
presentation, and title wording can be ignored.

\medskip
A SKU match additionally requires semantic agreement in
key sales specifications, including color, size, quantity,
and capacity.
A conflict in any key specification indicates a non-match.

\medskip
Use only the images, titles, and SKU attributes provided
in this request. Do not use external information.

\medskip
Options:

A. Match

B. Non-match

\medskip
You must strictly follow the JSON format below.
Do not include explanations or Markdown code blocks.

\medskip
\texttt{\{"answer": "A or B"\}}
\end{promptbox}

\end{document}